\documentclass[a4paper,10pt,twocolumn]{article}
\usepackage[english]{babel}
\usepackage[utf8]{inputenc}
\usepackage[T1]{fontenc}

\newcommand*{\tran}{^{\mkern-1.5mu\mathsf{T}}}
\usepackage[top=1.5cm, left=1.5cm, right=1.5cm, bottom=1.5cm]{geometry}

\renewenvironment{abstract}{\bf\small {\em\ Abstract---}}{}

\usepackage{amsfonts,amssymb,amsmath,amsthm}
\usepackage{graphicx}
\usepackage[footnotesize]{caption}

\usepackage{amsfonts,amssymb,amsmath,amsthm}

\usepackage{bm}

\usepackage{bbold} % for \mathbb{1}
\usepackage{color}
\usepackage{subcaption}

\title{Matrix Factorization via Deep Learning}

\author{Duc Minh Nguyen$^{1,2}$, Evaggelia Tsiligianni$^{1,2}$, Nikos Deligiannis$^{1,2}$. \\
\footnotesize $^1$Vrije Universiteit Brussel, Pleinlaan 2, B-1050 Brussels, Belgium \\
\footnotesize $^2$imec, Kapeldreef 75, B-3001 Leuven, Belgium \\
\footnotesize \{mdnguyen, etsiligi, ndeligia\}@etrovub.be
} 
\date{\empty}

\begin{document}

\maketitle

\begin{abstract} 
Matrix completion is one of the key problems 
in signal processing and machine learning. 
In recent years, deep-learning-based models have achieved state-of-the-art results 
in matrix completion. 
Nevertheless, they suffer from two drawbacks: 
(\textit{i}) they can not be extended easily to rows or columns unseen during training; 
and (\textit{ii}) their results are often degraded in case discrete predictions are required. 
This paper addresses these two drawbacks by presenting a deep matrix factorization model and 
a generic method to allow joint training of the factorization model and the discretization operator.
Experiments on a real movie rating dataset show the efficacy of the proposed models. 
\end{abstract}
\section{Introduction}
\label{sec:introduction}
Let $M \in \mathbb{R}^{n \times m}$ be an incomplete matrix
and $\Omega$ the set containing the indices of the observed entries in $M$ ($| \Omega | \ll n m $). 
Matrix completion (MC) concerns the task of recovering the unknown entries of $M$.
Existing work often relies on the assumption that $M$ is a low-rank matrix. 
Matrix factorization (MF) methods~\cite{Markovsky2012} 
approximate the unknown rank-$r$ matrix $M$ by the product of two factors, $UV^T$, 
$U \in \mathbb{R}^{n \times r}$, $V \in \mathbb{R}^{m \times r}$, and $r\ll \min(n,m)$.  

Recently, state-of-the-art performance 
in matrix completion has been achieved by neural network models ~\cite{sedhain15, zheng16, kuchaiev17}. 
Nevertheless, these methods suffer from two main drawbacks. 
The first is associated with their extendability to row or column samples unseen during training. 
The second concerns models that ignore the discrete nature of matrices involved in many application domains,
and produce sub-optimal solutions by applying a quantization operation (e.g.~rounding) 
on the real-valued predictions. 

In this work, we focus on these two drawbacks of deep-learning-based MC.
To address the first problem, we present a deep two-branch 
neural network model for MF, which can be effectively extended to rows and/or columns unseen during training. 
We obtain discrete predictions with a continuous approximation of a discretization operator, 
which enables simultaneous learning of the MF model and the discretizer. 
Experiments on a real dataset justify the effectiveness of our methods.  
\section{Deep Matrix Factorization Model}
\label{sec:dmf}
\begin{figure}[t]
	\centering
	\includegraphics[width=0.625\linewidth]{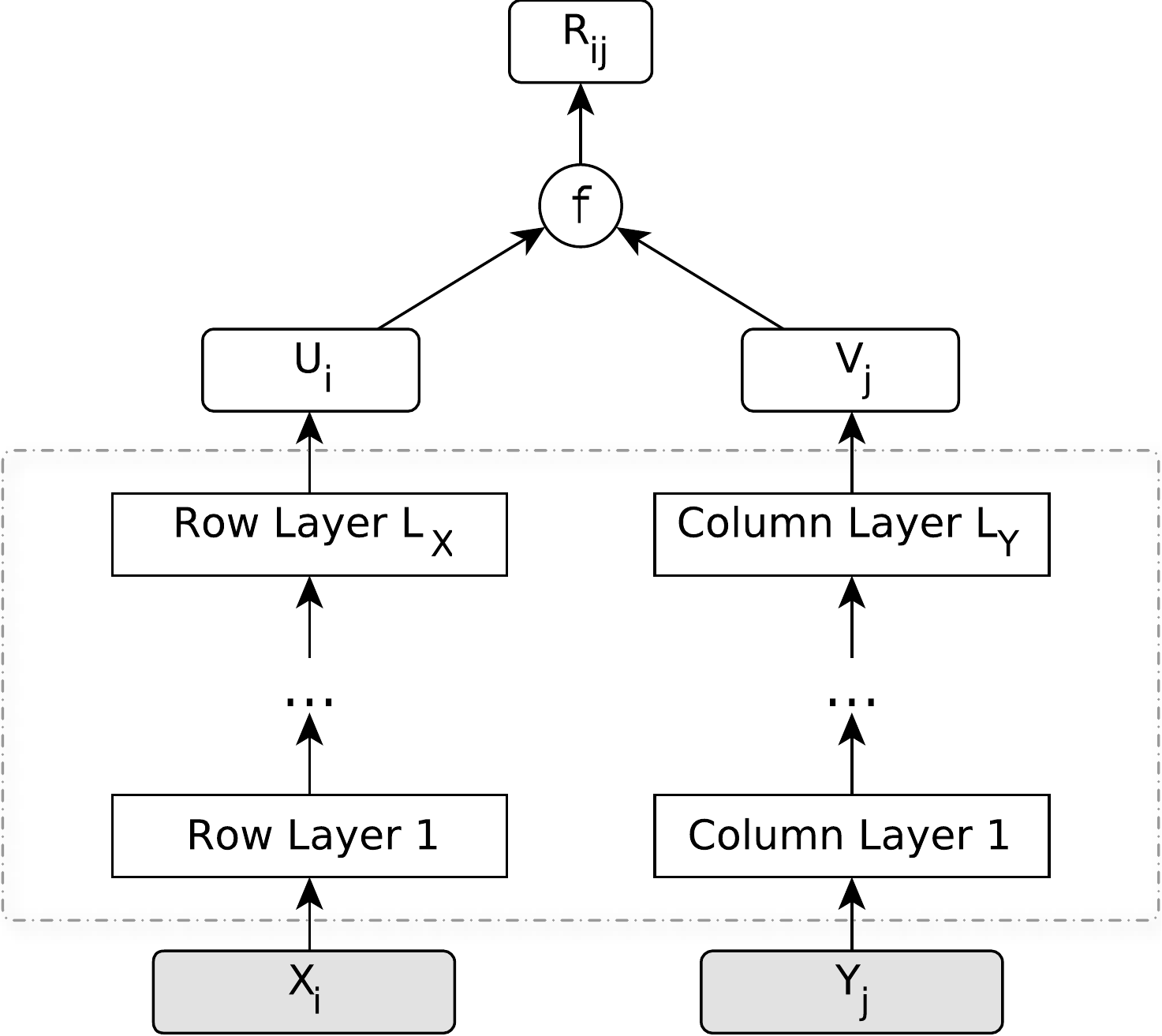}
	\caption{The deep matrix factorization model (DMF):
		The two branches consist of $L_X$ and $L_Y$ fully connected layers, 
		mapping the inputs $X_i, Y_j$ to the latent factors, $U_i$, $V_j$; 
		$f$ is the cosine similarity function.
	}
\label{fig:architecture}
\end{figure}

Consider a partially observed matrix $M \in \mathbb{R}^{n \times m}$,
and let $X_{i} \in \mathbb{R}^{m}, i = 1, \dots, n$, be the $i$-th row vector
and $Y_{j} \in \mathbb{R}^{n}, j = 1, \dots, m$, be the $j$-th column vector. 
Our model, illustrated in Fig.~\ref{fig:architecture}, comprises two branches 
that take as input the row and column vectors $X_i,Y_j$. 
The two branches can be seen as two embedding functions 
$h_X^{W_X}$ and $h_Y^{W_Y}$, realized by a number of $L_X$ and $L_Y$ 
fully connected layers, respectively; 
$W_X$ and $W_Y$ are the corresponding sets of weights. 
$h_X$ and $h_Y$ map $X_i$, $Y_j$ to the latent factors $U_i$, $V_j\in \mathbb{R}^{d}$. 
The prediction for the matrix entry at the $(i,j)$ position is then given by 
the cosine similarity between $U_i$ and $V_j$. 
%$f\left( U_i, V_j \right) = \frac{U_{i}^{T} V_{j}}{\left\|U_i \right\|_2 \left\|V_j \right\|_2}$. 
Therefore, our model maps the row and column vectors $X_i$, $Y_j$ to a continuous value  $R_{ij}  \in [-1,1]$\footnote{
During training, all entries $M_{ij} \in \left[ \alpha, \beta \right] $  are linearly scaled into $[-1,1]$, 
according to $M_{ij} = \frac{M_{ij} - \mu}{\mu - \alpha},~ \forall i,j$, 
with $\mu = \left( \alpha + \beta \right) / 2$.
A re-scaling step is required to bring the predicted values to the range $[\alpha, \beta]$.}
according to 
$R_{ij} = F(X_i,Y_j) = \frac{[h_X^{W_X}(X_i)]\tran  h_Y^{W_Y}(Y_j)}{\left\|h_X^{W_X}(X_i) \right\|_2 \left\|h_Y^{W_Y}(Y_j) \right\|_2}$.
We coin this model \textit{Deep Matrix Factorization} (DMF)\footnote{
DMF was presented in our previous work \cite{nguyen18}, and was independently proposed in \cite{xue17}.}.

We employ the mean square error as an objective function to train our model, with $\ell_2$ regularization 
on the network parameters $W_X, W_Y$. The final objective function has the form:
\begin{equation}
\label{eq:loss_dmf}
%\begin{aligned}
	\mathcal{L} = \dfrac{1}{|\Omega_{tr}|}  \sum_{ij \in \Omega_{tr}} \left(F(X_i,Y_j) - M_{ij} \right)^2 
	+ \gamma \left( \|W_X \|_{2}^{2} + \| W_Y \|_{2}^2 \right),
%\end{aligned}
\end{equation}
with $\gamma$ a hyperparameter.
\begin{figure}[t]
    \centering
    \includegraphics[scale=0.25]{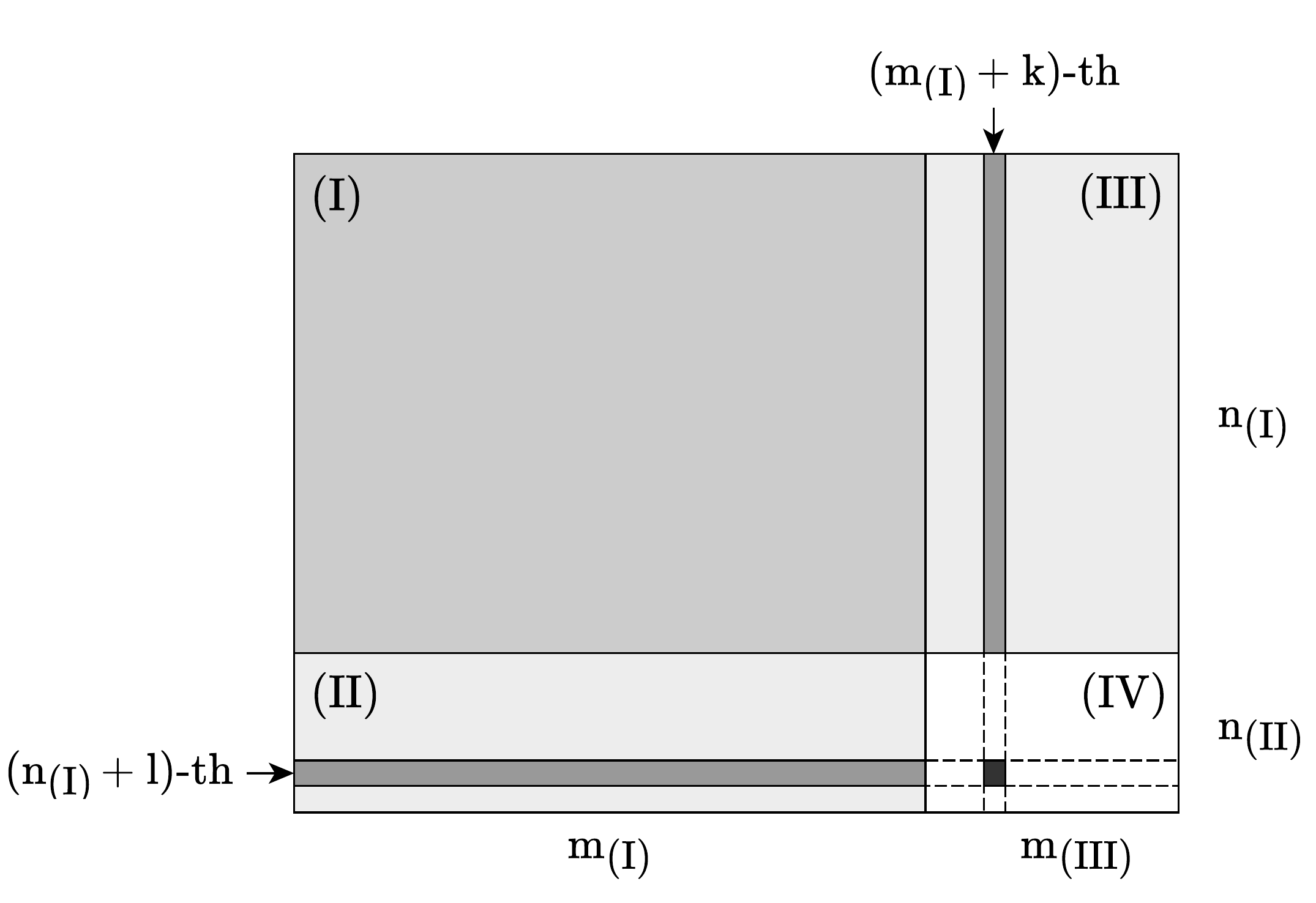}
    \caption{Extendability in matrix completion. 
    \textit{area} (I): rows and columns available during training.
    \textit{areas} (II) and (III): entries corresponding to the interactions of 
    unseen rows and seen columns and vice versa. 
    \textit{area} (IV): entries corresponding to the interactions of 
    unseen rows and unseen columns.}
\label{fig:rows_cols}
\end{figure}
\subsubsection*{Extendability of DMF}
An important question in MF is how to efficiently extend the model to samples unseen during training. 
For example, in recommender systems the trained model has to deal with new users and/or new items. 
Fig.~\ref{fig:rows_cols} illustrates different cases involving extendability. 
Incomplete rows and columns \textit{available} during training are depicted in Area (I); 
areas (II), (III) and (IV) represent new rows and columns which are only partially observed \textit{after} training. 
Let us denote by $M^{(I)}$, $M^{(I)\&(II)}$, $M^{(I)\&(III)}$ and $M^{(I)-(IV)}$ the matrices corresponding to the entries 
in areas (I), (I) and (II), (I) and (III) and (I), (II), (III), (IV), respectively. 
A DMF model trained with entries in $M^{(I)} \in \mathbb{R}^{n \times m}$, can be extended to other matrices as follows: 
\begin{align*}
	M^{(I)\&(II)}_{tj}   &= F \left( M^{(I)\&(II)}_{t,:}, M^{(I)}_{:,j} \right) \nonumber \\
	M^{(I)\&(III)}_{ik}  &= F \left( M^{(I)}_{i,:}, M^{(I)\&(III)}_{:,k} \right) \nonumber \\
	M^{(I)-(IV)}_{tk}    &= F \left( M^{(I)\&(II)}_{t,:}, M^{(I)\&(III)}_{:,k} \right),
\end{align*}
with $i=1,\dots,n$, $j=1,\dots,m$, $t > n$ and $k > m$. 
%
%%
%\begin{itemize}
%	\item{$M^{(I)\&(II)}_{tj}$ from $X_t=M^{(I)\&(II)}_{t,:}$, $Y_j=M^{(I)}_{:,j}$}
%	\item{$M^{(I)\&(III)}_{ik}$ from $X_i=M^{(I)}_{i,:}$, $Y_j=M^{(I)\&(III)}_{:,k}$}
%	\item{$M^{(I)-(IV)}_{tk}$ from $X_t=M^{(I)\&(II)}_{t,:}$, $Y_j=M^{(I)\&(III)}_{:,k}$}
%\end{itemize}
%%
%
\section{Discrete Matrix Factorization}
\label{sec:discrete}
Most matrix factorization models only produce real-valued entries 
from which discrete predictions can be obtained by a separate quantization step. 
Considering a set of quantization levels $\mathcal{I} = \{ I_1, \dots, I_d \}$, 
a quantizer $Q_{\boldsymbol{b}}$ divides the real line into $d$ non-overlapping consecutive intervals,
defined as the range $[b_{v-1}, b_v)$, $v=1, \dots, d$. 
A quantity will be mapped  to $I_v \in \mathcal{I}$ if it falls within the interval $[b_{v-1}, b_v)$~\cite{Gray1998}, that is,
\begin{equation}
   Q_{\boldsymbol{b}}(x) = \sum_{v=1}^d I_v \mathbb{1}_{b_{v-1} \leq x \leq b_v},
   \label{eq:quantization}
\end{equation}
where $\mathbb{1}_s$ outputs $1$ if the $s$ is true and $0$ otherwise,
and $\boldsymbol{b} = [b_0, \dots, b_d]^\intercal$ are the quantization boundaries.
%An example of a uniform quantizer with $d=5$ is illustrated in Fig.~\ref{fig:quantization}(a). 
In discrete MC, the set of quantization levels corresponds to the set of allowable discrete values of the matrix entries.

Denote by $H_{c}(x) \triangleq H(x-c)$ the heaviside step function, 
where $H(x-c) = 1$ if $x \geq c$ and $0$ otherwise.
Without loss of generality, let us assume that $I_{v+1}-I_{v} = \Delta$, for all $v=1,\dots, d - 1$.
We define $\boldsymbol{q} = [q_1,\  q_2, \ q_3, \dots , q_{d-1}, \ q_d] = [b_0, \ I_2, \ I_3, \dots, I_{d-1}, \ b_d]$.
Then, \eqref{eq:quantization} can be written as
%heaviside quantizer
\begin{equation}
	Q_{\boldsymbol{b}}(x) = \sum_{v=1}^{d-1} \Big[ I_{v}+ \Delta \cdot H_{b_{v}}(x) \Big] \mathbb{1}_{q_{v} \leq x < q_{v+1}}.
	\label{eq:quantization_explicit_compact}
\end{equation}
$Q_b$ is piecewise constant, thus, 
incorporating it into a gradient-based learning system such as DMF results in a vanishing gradient. 
We replace the heaviside step function $H_{b_{v}}$
by a logistic function of the form $\sigma_{\lambda, c} (x) = (1 + e^{-\lambda (x-c)})^{-1}$,
where $c$ is a scalar denoting the center of the sigmoid, 
and $\lambda$ controls the slope of $\sigma$. 
Therefore, we obtain the piecewise smooth function
\begin{equation}
	G_{\lambda, \boldsymbol{b}}(x) = \sum_{v=1}^{d-1} \Big[ I_{v}+ \Delta \cdot \sigma_{\lambda, b_{v}}(x) \Big] \mathbb{1}_{q_{v} \leq x < q_{v+1}}.
	\label{eq:smooth_quantization}
\end{equation}
%%%
Since $\lim_{\lambda \to +\infty} \sigma_c \left( \lambda, x \right) = H_c \left( x \right)$, 
$G_{\lambda, \boldsymbol{b}}$ becomes arbitrarily close to $Q_{\boldsymbol{b}}$ in \eqref{eq:quantization_explicit_compact},
when $\lambda$ becomes arbitrarily large.

By incorporating $G_{\lambda, \boldsymbol{b}}$ into the DMF model~(Sec.~\ref{sec:dmf}),
we obtain a discrete MF model coined DMF-D, which can be trained
using the following objective function:
\begin{align}
\label{eq:mc_mf_deep}
\mathcal{L} = \sum_{ij \in \Omega} & \left( G_{\lambda, \boldsymbol{b}} \left( F(X_i,Y_j) - M_{ij} \right) \right)^2 + \nonumber \\  
& \gamma_1 (\|W_X \|_{2}^{2} + \| W_Y \|_{2}^2) + \gamma_2 \|\boldsymbol{b} -\boldsymbol{\tilde{b}}\|_2^{2},
\end{align}
where the last term penalizes $G$ deviating significantly from a uniform quantizer, 
$\boldsymbol{\tilde{b}} = [\tilde{b}_0, \tilde{b}_1, \dots, \tilde{b}_d]$; 
$\gamma_1, \gamma_2$ are hyperparameters. 
We start with a small value of $\lambda$ at the beginning of training and gradually 
increase it to a very large value, 
so that the output of the model becomes discrete at the end of training. 
\section{Exprimental Results}
We carry out experiments on the MovieLens1M 
datasets \cite{harper15}, with 1 million user-movie ratings in $\{1,2,\dots,5\}$. 
We randomly split $75\%$ of the observed entries for training, $5\%$ for validation and $20\%$ for testing. 
We evaluate the prediction quality in terms of the root mean square error, 
$\text{RMSE} = \sqrt{\sum_{ij \in \Omega_\text{eval}} (R_{ij} - M_{ij})^2 / \left| \Omega_\text{eval} \right|}$, 
and the mean absolute error, 
$\text{MAE} = \sum_{ij \in \Omega_\text{eval}} \left| R_{ij} - M_{ij} \right| / \left| \Omega_\text{eval} \right|$, 
calculated over the entries reserved for testing ($\Omega_{eval}$). 
We report the results averaged over 5 runs with different random splits. 

Table~\ref{table:exp_1_ml1m} presents results regarding the extendability of 
different deep-learning-based models. Even though the DMF model does not 
produce the best predictions for entries in area (I), and (III), it produces the best results 
in area (II) and is the only model that can make predictions for entries in area (IV).  
\begin{table}
\centering
\caption{Results (RMSE) on the ML-1M dataset~\cite{harper15}, 
separated by areas (from (I) to (IV)).}
\label{table:exp_1_ml1m}
\begin{tabular}{ c | c | c | c | c }
\hline \hline
& (I) & (II) & (III) & (IV) \\
\hline
%U-CF-NADE-S \cite{zheng16} 			& $0.855$ & $0.671$ &    -    &    -    &    -    &    -    &    -    &    -    \\
%\hline
%I-CF-NADE-S \cite{zheng16} 			& $\textbf{0.839}$ & $\textbf{0.651}$ &    -    &    -    &    -    &    -    &    -    &    -    \\
%\hline
U-Autorec \cite{sedhain15}			& $0.906$ & $0.976$ &    -    &    -    \\
\hline
I-Autorec \cite{sedhain15}			& $0.841$ &    -    &  $\textbf{0.856}$ & -   \\
\hline
Deep U-Autorec \cite{kuchaiev17}	& $0.889$ & $0.969$ &    -    &     -    \\
\hline
DMF									& $0.850$ & $\textbf{0.883}$ & $0.864$ & $\textbf{0.904}$  \\
\hline \hline
\end{tabular}
\end{table}
\begin{table}[t]
\centering
\caption{Discrete MC results on the MovieLens1M dataset~\cite{harper15}.}
\label{table:exp_2_ml1m}
\begin{tabular}{ c | c | c }
\hline \hline
& RMSE & MAE \\
\hline
RankK \cite{huang13_ijcai} 		& $0.944$ & $0.656$ \\
\hline
OPTSPACE \cite{keshavan09}  	& $0.940$ & $0.665$ \\
\hline
RDMC \cite{huang13} 			& $0.987$ & $0.670$ \\
\hline
ODMC \cite{huo16}  				& $0.937$ & $0.658$ \\
\hline
DMF-D							& $\mathbf{0.898}$ & $\mathbf{0.625}$ \\
\hline \hline
\end{tabular} 
\end{table}
Table~\ref{table:exp_2_ml1m} presents a comparison of models that output \textit{discrete predictions}.
DMF-D outperforms existing models by large margins.
The results show the benefits of the proposed approximation 
and the join training of latent factors and the discretization operator. 
More experimental results can be found in~\cite{nguyen18,nguyen18_spl}.
\section{Conclusion}
\label{sec:conclusion}
We presented DMF, a deep neural network model for matrix factorization 
which can be extended to unseen samples without the need of re-training. 
We couple DMF with a method that allows to train discrete MF models with gradient descent, 
obtaining DMF-D, a strong model for discrete matrix completion. 
Experimental results with real data prove the effectiveness of both models 
compared to the state of the art. 
\bibliographystyle{plain}
\bibliography{refs_itwist18}

\end{document}